# Dynamic Matrix Decomposition for Action Recognition

## Abdul Basit

*University of Engineering and Technology, Peshawar*

---***---

**Abstract -** *Designing a technique for the automatic analysis of different actions in videos in order to detect the presence of interested activities is of high significance nowadays. In this paper, we explore a robust and dynamic appearance technique for the purpose of identifying different action activities. We also exploit a low-rank and structured sparse matrix decomposition (LSMD) method to better model these activities.. Our method is effective in encoding localized spatio-temporal features which enables the analysis of local motion taking place in the video. Our proposed model use adjacent frame differences as the input to the method thereby forcing it to capture the changes occurring in the video. The performance of our model is tested on a benchmark dataset in terms of detection accuracy. Results achieved with our model showed the promising capability of our model in detecting action activities.*

*Key Words*: Action activities, appearance features, motion trajectories.

## 1.INTRODUCTION

Nowadays video cameras can be easily installed for surveillance and monitoring crowded and complex environments including airport, public places, and other gathering halls. However, doing so is practically very challenging. Software and hardware systems for surveillance are often not smart due to limited number of trained personnel watching the scenes on monitor screens and the natural limitations of human attention capabilities. This is easy to image, when considering of the huge numbers of cameras that entail inspection, the monotonic nature of the surveillance videos, and the alertness entail to identify events and render fast response. It is worth noticing, even the seemingly easy task of searching achieved videos, off-line, for events that are known to have happened, needs the help of automated tools for video retrieval and summarization. Another aspect to the same matter is greater number of the people in the world living in cities [1][2][3]. Therefore, automated detection of action activities by self-organization phenomena resulting from the interactions of many individuals, can cause significant difficulty in this detection process [4]. It is significant to consider modern surveillance approaches, possibly in lieu of, or as an assistance to, human personnel taking care of monitoring process. There are still unimaginable problems due to the lack of empirical studies of action activities where apart from basic motion segmentation, also the analysis of unstructured behaviors, such as blocking, or unacceptable group interaction, is important for the safety of people during, for example, gatherings of public in different social gatherings, or in situations of sport excitement (winning of a sport). These situations can possibly cause action activity arising from the maximum density and irregular flow of people [5][6]. Moreover, the behavior of the people may transition from one state of collective behavior to a qualitatively different behavior depending on the current environment at hand. Such changes typically happen when people in the public gatherings accumulate, spread, or non-uniformly move with the crowd flow [7][8]. Activity analysis and scene perception need object detection, tracking and activity recognition [9]. These methods, entailing low-level motion features, appearance features, or group trajectories, depict good results in low level gathering of people, but fail in real-world high level gathering of people at different public places. State-of-the-art methods [10][11][12] show an offline data-driven approach for different videos to understand different action activities by considering long-term analysis. The method tracks people in these gatherings, presenting typical and rare behaviors. Other methods [13][14][15] introduce an interdisciplinary model for the analysis of the different action activities, which fuses strengths of simulation techniques, pedestrian detection and tracking, dense crowd detection and event detection [16][17].

Considering the challenges in detection and tracking in different action activities videos, the researchers have paid attention on collecting the motion information at a higher scale, thus not affiliating it to individual objects, but considering the activity as a single object [18][19][20][21]. These methods often entail low-level features such as multi-resolution histograms [22], spatio-temporal volumes [23][24], and appearance or motion descriptors [25][26][46]. In the work [27], the optical flow constraint is exploited to compute a conditional probability of the spatio-temporal intensity change. Furthermore, motion estimation and segmentation are fused into a functional minimization model based on a Bayesian technique. In [28][29], researchers exploited a mixture of dynamic textures to fit a video sequence and then provided homogeneous motion regions to the mixture components. However, the methods provided in [30][31][33] are only targeted at considering the cases of simple motion patterns. In [35][36], motion segmentation is carried out without relying on the optical flow. In [37][38][39][40], a dynamic texture method is proposed to compute the similarity between neighboring spatio-temporal proposals. These proposals are grouped by connected component analysis, resulting into over segmentation in presence of low level social gatherings. In [41][42][43][44][45], the researchers introduced a model to perform multi-target tracking in crowd using time integration of the dynamical system defined by the optical flow.

## 2. VIOLENT SOCIAL INTERACTION

In this paper, we explore an effective action activity detection method based on the work of [46] with an effective and adaptive appearance model. Within the proposed method, the combination of the generative technique and the discriminative classifier leads to a more flexible and reliable likelihood function to verify the state detections. Our proposed model is dynamically updated with consideration of occlusions to account for appearance variations and remove drifts.

Our proposed mode is designed within the Bayesian filtering framework (proposed by [46]) in which the goal is to find a posteriori probability, of the target state by

$$p(\mathbf{x}_t|\mathbf{z}_{1:t-1}) = \int p(\mathbf{x}_t|\mathbf{x}_{t-1})p(\mathbf{x}_{t-1}|\mathbf{z}_{1:t-1})d\mathbf{x}_{t-1},$$
$$p(\mathbf{x}_t|\mathbf{z}_{1:t}) \propto p(\mathbf{z}_t|\mathbf{x}_t)p(\mathbf{x}_t|\mathbf{z}_{1:t-1}),$$

In the formulation, the object state, and the observation at time t has been considered, where lx, ly, θ, s, α, φ present x, y translations, rotation angle, scale, aspect ratio, and skew respectively. It has been considered that the affine parameters are independent and designed by six scalar Gaussian distributions. The motion model presents that the state at t based on the immediate previous state, and the observation model shows the likelihood of observing zt at state xt. The particle filter is an effective way of Bayesian filtering, which computes the state regardless of the underlying distribution. The optimal state is found by the maximum a posteriori estimation over a set of N samples as equation in the equation below,

$$\hat{\mathbf{x}}_t = \arg\max_{\mathbf{x}_t^i} p(\mathbf{z}_t|\mathbf{x}_t^i)p(\mathbf{x}_t^i|\mathbf{x}_{t-1}),$$

Considering a particle filter, the samples at frame t can be determined by a Gaussian function with mean xt−1 and variance σ2 as provided in the formulation.

$$p(\mathbf{x}_t^i|\mathbf{x}_{t-1}) = G(\mathbf{x}_{t-1}, \sigma^2)$$

The exploitation of large number of samples would possibly ameliorate the action activities detection at the expense of increasing computational complexity. We fuse the detection result is the MAP estimation over the samples which can be formulated well with mode seeking, and propose a motion model. At time t, the sample set is obtained by the Gaussian distribution. The template set is consist of different tracking outcomes in the latest set of frames and the template in the first frame. Given the sample set, the sparse coefficients of each template set are computed according to the formulation.

$$\min_{\gamma_j} \|t^j - X\gamma_j\|_2^2 + \lambda_1\|\gamma_j\|_1, \quad \text{s.t. } \gamma_j \succeq 0, \quad j = 1,\ldots,m,$$

In the formulation, each column of X is a sample at time t and lambda is a weight parameter. The sample set X formulates an over-complete dictionary, and the sparsity constraints push to detect the samples that are highly in correspondence with the templates. The samples that do not model the templates well are not exploited as good candidates for the detection of action activities in the videos. We consider that this model is different from other methods which need solving 1-minimization problems. On the other side, our model entails solving 1-minimization problems, therefore minimizing the computational overheads to greater amount. Most methods use rectangular image regions to represent detection, and thus background pixels are inevitably included as part of the foreground parts. Therefore, classifiers based on local representations may be significantly affected when background blocks are exploited as positive ones for update. On the contrary, the holistic appearance captured by a target template is more powerful than the local appearance of local blocks. Therefore, holistic templates are more robust for discriminative methods to separate foreground parts from the background. Moreover, local representations are more effective for generative methods due to their flexibility. In our proposed model, we show a collaborative observation model that integrates a discriminative classifier based on holistic templates and a generative model considering local representations.

We also explored the method of Peng et al. [47] who introduced a low-rank and structured sparse matrix decomposition (LSMD) method. For this purpose, a tree-structured sparsity-inducing norm regularization is firstly proposed to present a hierarchical description of the image structure to ensure the completeness of the extracted features. The similarity of feature values within the action activities detection is then provided by the `1-norm. High-level priors are put together to help the matrix decomposition and improve the detection of activities.

Tree structure is widely existed and exploited in natural image processing, e.g., tree-structured wavelet transforms, tree-based image segmentation. Recent advances in the sparse representation research also use tree structure to pursuit the structured sparsity in terms of relationships between different motion patterns. The paper uses a tree structured sparsity-inducing norm, which definitely is a hierarchical group sparsity, to represent the underlying structure of activities in feature space.

The stage is to formulate an index tree to model the scene structure via divisive hierarchical k-means clustering. During the tree modeling, for each image proposal, we receive its position coordinate and feature representation. All proposals from an image composite a set of different data points. The divisive hierarchical clustering starts with all the points in a single cluster, and then recursively categorizes each cluster into k child clusters using k-means method. The recursion ends when all the clusters consist less than k data points. We use a quad-tree structure where k = 4.

After completing the feature matrix representation and the corresponding structured index tree of the input scene, we explore the proposed LSMD method to categorize it into a low-rank part and a structured sparse part. Using the tree-structured sparsity regularization into the LSMD technique, we can fuse perceptually similar proposals of the foreground parts and removing the irrelevant pixels.

## 3. EXPERIMENTS

Our proposed model executes on a Windows system. Eight different people were tested performing various kinds of action activities in a set of videos [48]. Dataset was gathered with a Sony stationary camera. The experimental results gathered with our model are presented in the table below. Our proposed model significantly detected most of the action activities. It presents the strong capability of our proposed model to detect different action activities in videos.

| Name | Total Frames | No. of action activities | Correct detection |
|---|---|---|---|
| Omar Yun1 | 510 | 6 | 4 |
| Omar Yun2 | 279 | 1 | 1 |
| Yun Omar1 | 400 | 2 | 1 |
| Yun Omar2 | 239 | 1 | 0 |
| Zeeshan Cen1 | 474 | 4 | 3 |
| Zeeshan Cen2 | 333 | 4 | 3 |
| Jaime Jigna1 | 329 | 2 | 2 |
| Jaime Xuan2 | 408 | 3 | 2 |
| Joanna Xu1 | 281 | 4 | 3 |
| Joey Dave1 | 310 | 7 | 6 |
| Joey Dave2 | 249 | 2 | 1 |
| Dave Will 1 | 130 | 1 | 1 |
| Joey Joanna | 210 | 2 | 1 |
| Kris Rusty1 | 465 | 6 | 3 |
| Kris Rusty2 | 304 | 2 | 1 |

## 3. CONCLUSIONS

In this paper, we proposed a robust model to detect the presence of different action activities in videos. For this purpose, we exploited a robust appearance model with the addition of a low-rank and structured sparse matrix decomposition (LSMD) model to better preseent the existence of action activities. Our proposed method is capable of encoding spatio-temporal features which enables the analysis of local motion taking place in the video. The performance of our method is evaluated on a standard dataset. The experimental results of our proposed model show the promising capability of our method in detecting different action activities.